\documentclass[11pt]{article}
\usepackage{url,hyperref,lineno,microtype,subcaption}
\usepackage[onehalfspacing]{setspace}
\usepackage{amsmath}
\usepackage{amsfonts}
\usepackage{mathrsfs}
\usepackage{bbm}

\def\bbR{{\mathbb{R}}}

\def\calN{{\mathcal{N}}}
\def\calO{{\mathcal{O}}}

\def\H{{\mathbf{H}}}
\def\I{{\mathbf{I}}}

\def\Q{{\mathbf{Q}}}

\def\a{{\mathbf{a}}}

\def\e{{\mathbf{e}}}

\def\g{{\mathbf{g}}}

\def\u{{\mathbf{u}}}

\def\w{{\mathbf{w}}}
\def\x{{\mathbf{x}}}
\def\y{{\mathbf{y}}}

\usepackage{amsthm}
\usepackage{etoolbox}
\theoremstyle{plain}

\usepackage{multirow}
\usepackage{graphicx}
\usepackage{pifont}
\usepackage{tabularx}
\usepackage{array}
\usepackage{makecell}
\usepackage{authblk}
\usepackage[numbers]{natbib}
\usepackage{booktabs} 
\usepackage{caption}  
\newcolumntype{C}{>{\centering\arraybackslash}X}
\DeclareMathOperator*{\argmax}{arg\,max}

\begin{document}
\title{Electroencephalogram Emotion Recognition \\ via AUC Maximization}

\author[]{Minheng Xiao$^1$\footnote{Corresponding author: minhengxiao@gmail.com}~}
\affil[]{
$^1$Department of Integrated System Engineering, Ohio State University, Columbus, OH, USA\\
$^2$Department of Mathematics and Statistics, Boston University, Boston, MA, USA}

\date{} 

\maketitle
\begin{abstract}
Imbalanced datasets pose significant challenges in areas including neuroscience, cognitive science, and medical diagnostics, where accurately detecting minority classes is essential for robust model performance. This study addresses the issue of class imbalance, using the `Liking' label in the DEAP dataset as an example. Such imbalances are often overlooked by prior research, which typically focuses on the more balanced arousal and valence labels and predominantly uses accuracy metrics to measure model performance. To tackle this issue, we adopt numerical optimization techniques aimed at maximizing the area under the curve (AUC), thus enhancing the detection of underrepresented classes. Our approach, which begins with a linear classifier, is compared against traditional linear classifiers, including logistic regression and support vector machines (SVM). Our method significantly outperforms these models, increasing recall from 41.6\% to 79.7\% and improving the F1-score from 0.506 to 0.632. These results highlight the efficacy of AUC maximization via numerical optimization in managing imbalanced datasets, providing an effective solution for enhancing predictive accuracy in detecting minority but crucial classes in out-of-sample datasets.

\end{abstract}

\section{Introduction}
\label{sec: Intro}
Electroencephalography (EEG) plays a critical role in the study of human brain function~\citep{foreman2012quantitative}, offering valuable insights into cognitive~\citep{chikhi2022eeg}, emotional~\citep{li2022eeg}, and motor processes~\citep{al2021deep}. The ability to measure the brain’s electrical activity in real-time with high temporal resolution makes EEG an indispensable tool in neuroscience and clinical diagnostics. Its application extends from understanding basic cognitive functions to complex tasks like emotion recognition, where EEG provides a direct window into the brain's response to various stimuli. Unlike other physiological signals, EEG is highly sensitive to the brain's intrinsic activities, offering a unique perspective on neural dynamics that cannot be easily captured through other modalities~\citep{zhang2023applied, lim2024review}.

In recent years, EEG-based emotion recognition has gained considerable attention, particularly in the fields of brain-computer interfacing and human-machine interaction~\citep{liu2024distributed, cao2017structurally, al2017review}. Various methods have been developed to decode emotional states from EEG signals, leveraging the rich temporal and spatial information embedded in these signals~\citep{li2022multi}. Traditional machine learning approaches, such as support vector machines (SVM) and k-nearest neighbors (KNN), have been widely used, often with features like wavelet transforms and power spectral density~\citep{amin2015feature, bazgir2018emotion}. More recently, deep learning models, including convolutional neural networks (CNNs) and long short-term memory (LSTM) networks, have shown promise in improving classification accuracy by automatically extracting complex features from raw EEG data~\citep{li2022deep, walther2023systematic}. These advancements highlight the potential of EEG in emotion recognition.

Despite these advancements, a significant challenge in EEG-based emotion recognition is the issue of class imbalance~\citep{houssein2022human}. In many EEG datasets, the distribution of emotional states is skewed, with some emotions being underrepresented. This imbalance can lead to biased classifiers that perform well on majority classes but fail to accurately detect minority classes. Various strategies have been developed to address this issue, including data augmentation techniques like synthetic minority over-sampling technique (SMOTE) and generative adversarial network based approaches~\citep{lashgari2020data, chen2021effects}. These methods aim to artificially balance the dataset by generating synthetic samples for the minority classes, thereby improving the model's ability to generalize across different emotional states. However, these solutions often involve complex preprocessing steps and can introduce noise, potentially affecting the overall model performance.

In this paper, we aim to tackle the class imbalance problem in EEG-based emotion recognition using a more straightforward and effective approach: Area under the Curve (AUC) Maximization~\citep{yang2022auc}. By focusing on maximizing the AUC with a simple linear classifier, we propose a method that directly addresses the imbalanced nature of EEG datasets. This approach not only simplifies the model training process but also enhances the detection of minority classes, leading to more robust and reliable emotion recognition systems. Through extensive experiments, we demonstrate the effectiveness of our method on standard EEG datasets, showing significant improvements in classification performance, particularly for underrepresented emotional states.

In Section~\ref{sec: Related}, we review related works and provide a summary table that includes models, methods, and results from these studies. In Section~\ref{sec: Dataset}, we introduce the DEAP dataset and discuss the preprocessing and feature engineering involved. In Section~\ref{sec: AUC}, we introduce several first-order and second-order algorithms for addressing AUC maximization problems. In Section~\ref{sec: Experiment}, we perform numerical experiments on the preprocessed and feature-engineered DEAP dataset on four feature sets for binary classification tasks.

\section{Related Works}
\label{sec: Related}
Recent advancements in EEG-based emotion classification on DEAP dataset have demonstrated significant progress, moving from simple traditional models to more complex deep learning models. Initially, studies utilized relatively straightforward approaches. For instance, Ali et al.~\citep{ali2016eeg} employed a combination of wavelet energy, modified energy, wavelet entropy, and statistical features with SVM, achieving accuracies of 84.95\% for valence and 84.14\% for arousal. Meanwhile, Li et al.~\citep{li2016emotion} made contributions in feature engineering by using wavelet transforms and scalogram to structure multichannel neurophysiological signals. This method, combined with a hybrid deep learning model of CNNs and recurrent neural networks (RNNs), achieved accuracies of 72.06\% for valence and 74.12\% for arousal.

As technology advances, deep learning models such as deep neural networks (DNNs)~\citep{jiang2021recurrent, liufu2024neural} and CNNs~\citep{liu2024infrared, liufu2021reformative} become more prevalent. Tripathi et al.~\citep{tripathi2017using} utilize DNNs and CNNs to classify emotions from the DEAP dataset, achieving accuracy of 81.41\% and 73.36\% in accuracy for valence and arousal, respectively. Cheah et al.~\citep{cheah2019short} develop user-specific CNN models, achieving remarkable accuracies of 98.75\% for valence and 97.58\% for arousal in a three-class classification task.

Some studies explore more complex deep learning architectures and optimization techniques~\citep{wei2024collaborative, liufu2022modified}. For instance, Liu et al.~\citep{liu2016emotion} leverage a Bimodal Deep AutoEncoder (BDAE), outperforming existing methods with an average accuracy of 83.25\% on multimodal tasks using the DEAP dataset. Gao et al.~\citep{gao2020channel} introduce the Channel-Fused Dense Convolutional Network (CDCN), significantly improving accuracy to 92.24\% for valence and 92.92\% for arousal. Kumar et al.~\citep{kumar2022human} apply CNN and multilayer perceptron (MLP) models, achieving F1-scores of 94.5\% and 94\% for arousal and valence classes, respectively.

Additionally, Zhang et al.~\citep{zhang2023attention} develop an attention-based hybrid model, which achieves accuracies of 85.86\% for arousal and 84.27\% for valence. Meanwhile, Zheng et al.~\citep{zheng2023adaptive} introduce an Adaptive Neural Decision Tree (ANT), achieving an impressive 99.14\% average accuracy in binary classification tasks.

\textbf{Comparison to Prior Works}: Compared to the extensive prior research on the DEAP dataset (See Table~\ref{tab: related works} for a limited reference), which predominantly focuses on the accuracy of models in the two main categories of arousal and valence, our work targets the `Liking' category. This category is notably characterized by an imbalanced data distribution (See Figure~\ref{fig: labels}). Imbalance issues are observed in various domains, including cognitive science, where certain cognitive states may be less frequently represented, and in medical diagnosis, where certain diseases or conditions may be rare. For example, in medical imaging, datasets often contain a majority of healthy images with only a small fraction depicting abnormalities, leading to challenges in training robust models~\citep{lin2020touch}. Despite the prevalence of such imbalances, many researchers continue to rely on machine learning methods trained on artificially balanced datasets. While this approach may yield high in-sample performance, it often results in poor out-of-sample generalization when the data distribution differs from the training set. Given that the `Liking' category is imbalanced, our work primarily addresses this issue using numerical methods, specifically AUC maximization, to improve the detection accuracy of minority yet crucial classes. This approach is aimed at enhancing the precision of detecting these underrepresented but important categories, thereby ensuring better real-world applicability.

\begin{table}[htbp]
\centering
\caption{Summary of Methods and Results on the DEAP Dataset}
\begin{tabular}{@{}p{1.5cm}p{8cm}p{6cm}@{}}
\toprule
\textbf{Article} & \textbf{Existing Methods} & \textbf{Results} \\ 
\midrule
\cite{zheng2023adaptive} & Adaptive Neural Decision Tree & Average accuracy 99.14\% (Binary) \\
\hline
\addlinespace
\cite{iyer2023cnn} & Ensemble Learning (CNN + LSTM) & Average accuracy 65\% \\
\hline
\addlinespace
\cite{zhang2023attention} & Attention-based Hybrid Deep Learning & Accuracy 85.86\% for arousal \newline Accuracy 84.27\% for valence \\
\hline
\addlinespace
\cite{lin2023eeg} & Improved Graph Neural Network & Accuracy 90.74\% (DEAP-Twente) \newline Accuracy 91\% (DEAP-Geneve) \\
\hline
\addlinespace
\cite{wang2023multimodal} & Multimodal Emotion Recognition & Accuracy 96.63\% for valence \newline Accuracy 97.15\% for arousal \\
\hline
\addlinespace
\cite{kumar2022human} & CNN and MLP & 94.5\% F1-score for arousal \newline 95\% F1-score for valence \\
\hline
\addlinespace
\cite{cimtay2020investigating} & InceptionResnetV2 & Average accuracy 72.81\% (Binary) \\
\hline
\addlinespace
\cite{abdel2020emotion} & Log-Euclidean Riemannian Metric & Accuracy 72.6\% for arousal \newline Accuracy 74.6\% for valence \\
\hline
\addlinespace
\cite{luo2020eeg} & Spiking Neural Networks (SNNs) & Accuracy (arousal, 74\%) \newline Accuracy (valence, 78\%) \newline Accuracy (dominance, 80\%) \newline Accuracy (liking, 86.27\%) \\
\hline
\addlinespace
\cite{gao2020channel} & Channel-Fused Dense Convolutional Network (CDCN) & Accuracy 92.24\% for valence \newline Accuracy 92.92\% for arousal \\
\hline
\addlinespace
\cite{sharma2020automated} & LSTM with Higher-Order Statistics & Average accuracy 82.01\% \\
\hline
\addlinespace
\cite{cheah2019short} & CNN & Around 98\% accuracy for 3-class arousal and valence \\
\hline
\addlinespace
\cite{tripathi2017using} & CNN & Accuracy 81.41\% for valence \newline Accuracy 73.36\% for arousal \\
\hline
\addlinespace
\cite{li2016emotion} & C-RNN & Accuracy 72.06\% for valence \newline Accuracy 74.12\% for arousal \\
\hline
\addlinespace
\cite{liu2016emotion} & MDL & Accuracy 85.2\% for valence \newline Accuracy 80.5\% for arousal \\
\hline
\addlinespace
\cite{ali2016eeg} & WT-SVM & Accuracy 84.95\% for valence \newline Accuracy 84.14\% for arousal \\
\bottomrule
\end{tabular}
\label{tab: related works}
\end{table}

\section{Dataset}
\label{sec: Dataset}
In this section, we introduce the DEAP dataset~\citep{koelstra2011deap, scherer2005emotions} (Section~\ref{subsec: Dataset/overview}) along with the feature engineering procedures employed (Section~\ref{subsec: Dataset/FE}). The feature engineering is divided into three types: frequency domain, time domain, and frequency-time domain. A summary of the feature engineering processes is presented in Table~\ref{tab: feature engineering}.

\subsection{Overview}
\label{subsec: Dataset/overview}
The DEAP dataset is widely used in studying the relationship between physiological signals and emotional states. The dataset consists of data from 32 participants who each watch 40 one-minute video clips, selected from an initial pool of 120 based on subjective emotional assessments. EEG signals are recorded using a Biosemi ActiveTwo system with 32 AgCl electrodes, following the international 10-20 system. Each participant's EEG data is captured across 40 channels, originally sampled at 512 Hz and later downsampled to 128 Hz. The data also undergoes preprocessing, including the removal of Electrooculography (EOG) artifacts and band-pass filtering between 4.0 and 45.0 Hz. Additionally, each video trial includes a 3-second pre-trial baseline, with the total recording per trial amounting to 63 seconds.

\begin{figure}[htbp]
\centering
\includegraphics[scale=0.65]{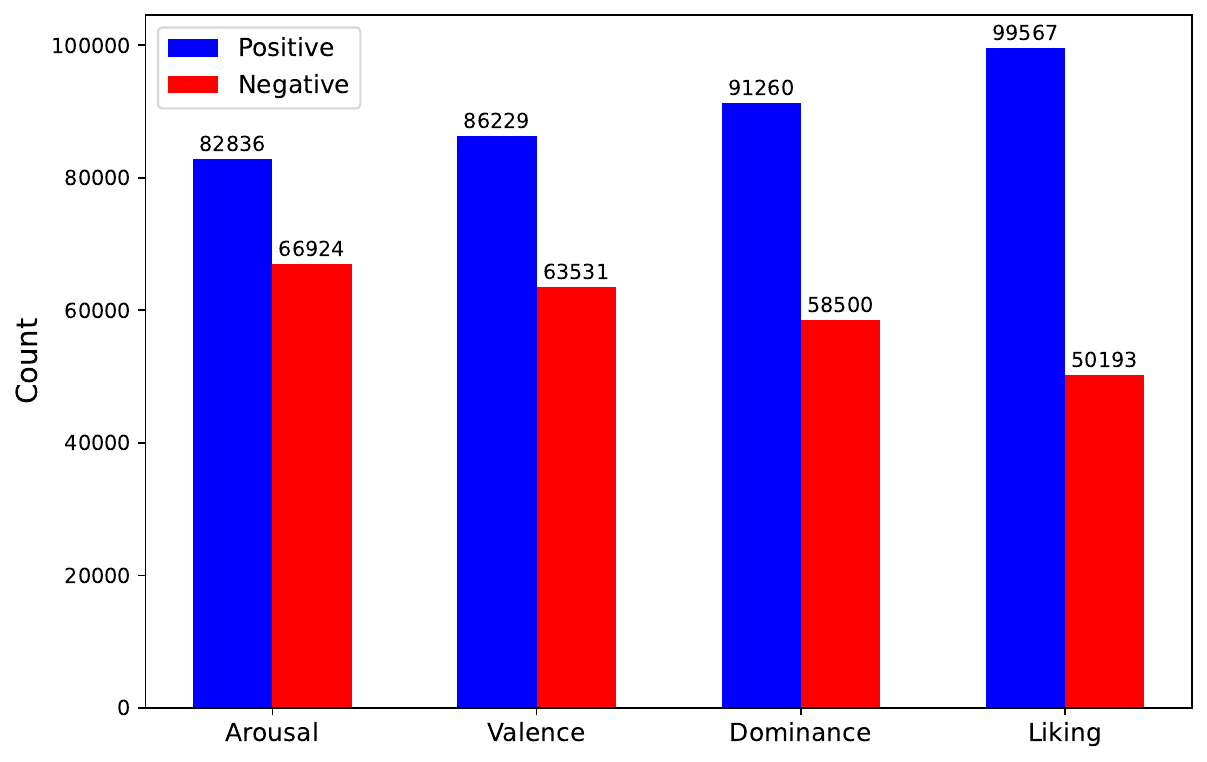}
\caption{Class distribution for the DEAP dataset across four emotional categories. Notably, the `Liking' label exhibits a significant imbalance, with a 2:1 ratio between the two classes.}
\label{fig: labels}
\end{figure}

The dataset includes four primary emotion labels: Arousal, Valence, Dominance, and Liking. Among these labels, arousal and valence categories are commonly used due to their relatively balanced distributions, facilitating effective model evaluation in machine learning and deep learning applications. However, the `Liking' label is notably imbalanced (See Figure~\ref{fig: labels}), with approximately one-third of scores below 5 and two-thirds above, presenting challenges for accurate model assessment. This kind of label imbalance is typical in neuroscience and medical datasets, such as those related to Epilepsy, Dementia, Parkinson's disease, and Depression~\citep{vuttipittayamongkol2020improved, aouto2023comprehensive, edward2023new}.

A summary of the DEAP dataset and the channels utilized in our study can be seen in Table~\ref{tab:deap_summary}. Given the imbalanced nature of certain features and their significance in detecting diseases, particularly those with fewer samples, it is crucial to develop methods that can effectively identify these cases within large datasets. Our objective is to employ advanced numerical optimization techniques to address these challenges and improve the predictive accuracy and reliability of models handling imbalanced datasets, thereby providing more robust tools for research and clinical applications in neuroscience and health.

\begin{table}[htbp]
\centering
\caption{Summary of DEAP Dataset and Channels Used for Feature Extraction}
\begin{tabularx}{\textwidth}{@{}lX@{}}
\toprule
\textbf{Name} & \textbf{Description} \\
\midrule
Original Data & $40\times40\times8064$ (\#videos$\times$\#channels$\times$\#samples) per subject, 32 subjects total \\
\addlinespace
Label & $40 \times 4$ (\#videos $\times$ \#labels) per subject; labels: arousal, valence, dominance, liking \\
\addlinespace
Channel Index & Channels [1, 2, 3, 4, 6, 11, 13, 17, 19, 20, 21, 25, 29, 31] used for feature extraction \\
\addlinespace
Sampling Frequency & 128 Hz \\
\addlinespace
Window Size & 2-second for segments, 0.5-second for sliding \\
\addlinespace
Continents & Theta (4–8 Hz), Alpha (8–13 Hz), Beta 
 (13–30 Hz), and Gamma (30–45 Hz) \\
\bottomrule
\end{tabularx}
\label{tab:deap_summary}
\end{table}

\subsection{Feature Extraction}
\label{subsec: Dataset/FE}

\subsubsection{Sliding Window Segments}
\label{subsubsec: Dataset/FE/Sliding}
\begin{figure}[htbp]
\centering
\includegraphics[scale=0.7]{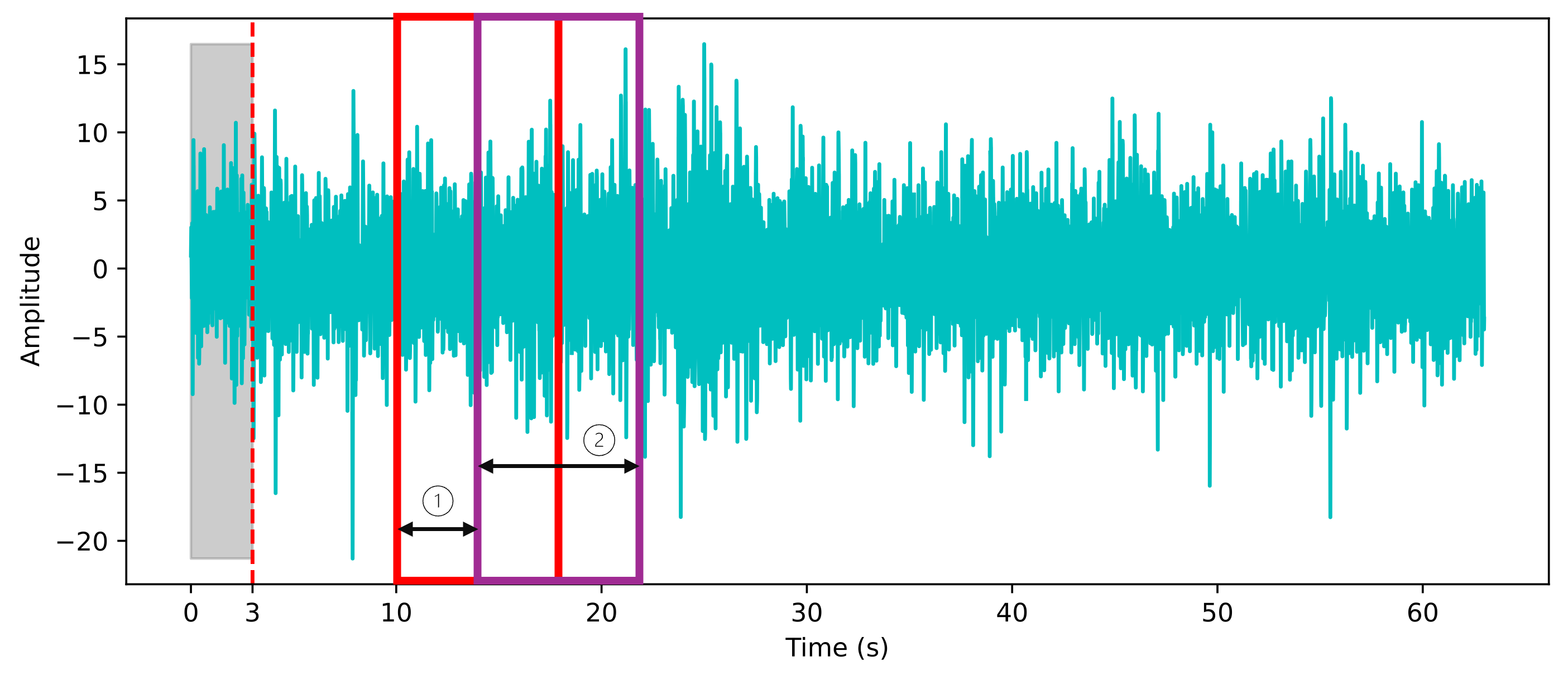}
\caption{An example of a DEAP EEG channel sliding window segmentation. The channel is 63 seconds long, with a 3-second pre-trial period denoted by the grey area. The red and purple rectangles represent a segment with a length denoted by circle 2, while the sliding window size is indicated by circle 1.}
\label{fig: channel segment}
\end{figure}

We first remove the initial 3-second pre-trial signals (the grey area in Figure~\ref{fig: channel segment}). To segment the continuous EEG signal into epochs, we employed a sliding window technique. Each channel is divided into consecutive 2-second segments, with the window sliding forward by 0.5 seconds for each new segment. This method results in overlapping segments, allowing for a more detailed temporal analysis of the signal. Given the sampling rate of 128 Hz, each 2-second segment consists of 256 data points. Consequently, a total of 477 distinct segments (epochs) are generated for each EEG channel. This sliding window approach facilitates the extraction of time-series signals from a single channel, enabling the study of variations over time.

\subsubsection{Frequency-domain Features}
\label{subsubsec: Dataset/FE/Freq}
One of the techniques commonly applied to extract frequency-domain features from EEG signals is the Fast Fourier Transform (FFT). FFT is a computationally efficient algorithm that transforms a signal from its time domain into its frequency domain, allowing for the analysis of the signal's frequency components. Given a discrete time-domain signal $x[n]$ of length $N$, the Discrete Fourier Transform (DFT) of the signal can also be expressed using Euler's formula as
\begin{align*}
X[k] = \sum_{n=0}^{N-1} x[n] \cdot \left[\cos\left(\frac{2\pi}{N} kn\right) - j\sin\left(\frac{2\pi}{N} kn\right)\right], \quad k = 0, 1, \dots, N-1
\end{align*}
Here, the DFT is decomposed into its real part, represented by the cosine term, and its imaginary part, represented by the sine term. The FFT is an efficient implementation of the DFT, reducing the computational complexity from $O(N^2)$ to $O(N \log N)$, making it particularly suitable for analyzing large datasets like EEG signals. By applying the FFT, the EEG signal can be decomposed into its constituent frequencies, including theta (4–8 Hz), alpha (8–13 Hz), beta (13–30 Hz), and gamma (30–45 Hz). Subsequently, the Power Spectral Density (PSD) is calculated for each frequency band within each sliding window segment. The PSD quantifies the power distribution of the signal over the different frequency components, providing insights into the underlying neural oscillations within each epoch. This feature is crucial for characterizing the energy present in specific frequency bands, which is often associated with various cognitive and emotional states~\citep{kim2018effective, qin2019extract, rahman2021emotion}.

\subsubsection{Frequency/Time-domain Features}
\label{subsubsec: Dataset/FE/FreqTime}
Instead of directly converting the signal from the time-domain to the frequency-domain, it is also promising to consider the frequency/time-domain features. Specifically, given the channel in a time-series manner, we adopt a Butterworth filter to extract the theta, alpha, beta, and gamma constituents of different frequencies while staying in the time domain. The Butterworth filter features a smooth frequency response in the passband and a seamless transition to the stopband. Mathematically, the magnitude response of an $n$-th order Butterworth filter is given by
\begin{align*}
|H(\omega)|^2 = \frac{1}{1 + \left(\frac{\omega}{\omega_c}\right)^{2n}},
\end{align*}
where $\omega_c$ is the cutoff frequency and $n$ is the order of the filter. When the cutoff frequency is normalized to $\omega = 1$ radian per second, the expression simplifies to:
\begin{align*}
|H(\omega)|^2 = \frac{1}{1 + \omega^{2n}}.
\end{align*}
Within each sliding window segment, the differential entropy (DE) features from each frequency band is extracted, which has been proved to be the most stable feature for emotion recognition~\cite{du2023eeg, lu2023emotion}. Morever, the statistical features are also extracted for each constituents.

\subsubsection{Cross-Channel Features}
\label{subsubsec: Dataset/FE/Cross}
\begin{figure}[htbp]
\centering
\includegraphics[scale=0.7]{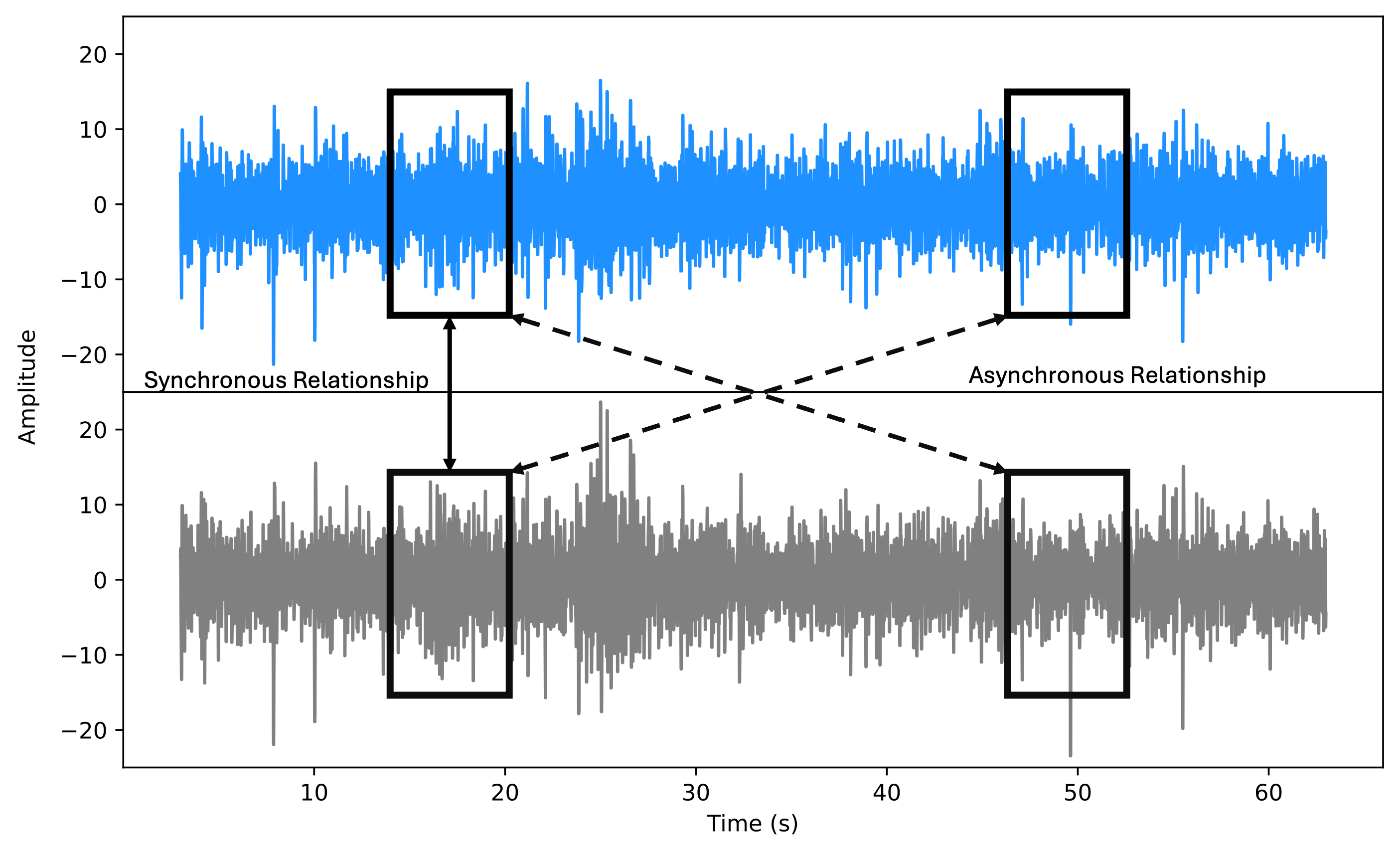}
\caption{An illustration of cross-channel features, showing the PLV and correlation between segments across various channels in both synchronous and asynchronous fashions.}
\label{fig: cross-channel feature}
\end{figure}

Besides analyzing a single channel in the frequency and time domains, it is also valuable to consider the relationships between channels, as these interactions may be key to emotion recognition (see Figure~\ref{fig: cross-channel feature}). Specifically, we focus on the Phase Locking Value (PLV) and the correlation between two segments in both synchronous and asynchronous fashions. The PLV measures the phase synchrony between signals, indicating how consistently two channels are phase-locked over time. Mathematically, the PLV between two segments $x(t)$ and $y(t)$ is defined as
\begin{align*}
\text{PLV} = \left| \frac{1}{N} \sum_{n=1}^{N} e^{i\Delta\phi_n(t)} \right|,
\end{align*}
where $\Delta\phi_n(t) = \phi_x(t) - \phi_y(t)$ is the phase difference between two signals at time $t$. The Pearson coefficient $r_{xy}(\tau)$ between two segments $x(t)$ and $y(t)$ can be computed with a time lag $\tau$ as
\begin{align*}
r_{xy}(\tau) = \frac{\sum_{t=1}^{T-\tau} (x(t) - \bar{x})(y(t+\tau) - \bar{y})}{\sqrt{\sum_{t=1}^{T-\tau} (x(t) - \bar{x})^2 \sum_{t=1}^{T-\tau} (y(t+\tau) - \bar{y})^2}},
\end{align*}
where $\bar{x}$ and $\bar{y}$ are the means of the segments. For synchronous cases, this reduces to $r_{xy}(0)$, where $\tau = 0$.

\begin{table}[htbp]
\centering
\caption{A summary of feature extraction methods for EEG signals. F represents Frequency domain features, T represents Time domain features, and FT represents Frequency-Time domain features.}
\resizebox{\textwidth}{!}{
\begin{tabular}{cccc}
\hline
\addlinespace[2pt]
Type & Feature & Sliding? & Description \\
\hline
\addlinespace[2pt]
F & PSD & $\boldsymbol{\checkmark}$ & Power Spectral Density, representing power distribution over frequency bands \\
\hline 
\addlinespace[2pt]
\multirow{2}{*}{T} & PLV & $\boldsymbol{\checkmark}$ & Phase Locking Value, measuring phase synchronization between channels \\
 & Corr & $\boldsymbol{\checkmark}$ & Correlation between time-series segments across channels \\
\hline
\addlinespace[2pt]
\multirow{6}{*}{FT} & DE & $\boldsymbol{\checkmark}$ & Differential Entropy, measuring signal complexity \\
 & Segment Range & $\boldsymbol{\checkmark}$ & Range of values within a segment, capturing signal variability \\
 & Segment Moments & $\boldsymbol{\checkmark}$ & Statistical moments (mean, variance, etc.) of the segment \\
 & Segment Diff & $\boldsymbol{\checkmark}$ & Difference between two consecutive segments \\
 & Channel Range & \textbf{\ding{55}} & Range of values of channels \\
 & Channel Moments & \textbf{\ding{55}} & Statistical moments (mean, variance, argmin, argmax, etc.) of channels \\
\hline
\end{tabular}
}
\label{tab: feature engineering}
\end{table}

\section{AUC Maximization Algorithms}
\label{sec: AUC}
The AUC metric evaluates classification model performance by measuring the area under the receiver operating characteristic (ROC) curve, reflecting the model's ability to differentiate between classes at various thresholds. Unlike traditional accuracy, which measures correct predictions irrespective of class distribution, AUC provides a comprehensive assessment across all thresholds, making it especially useful in imbalanced class scenarios. Most algorithms do not directly optimize AUC due to its non-decomposable and threshold-independent nature, using it instead as a post-training performance metric. However, AUC maximization can be approached through a strongly-convex-strongly-concave minimax optimization~\citep{liu2022quasi, liu2022partial}, enabling direct optimization and offering a robust alternative to conventional loss functions focused on point estimates. In order to train a linear classifier $\w$, the objective function can be written as
\begin{align}
\label{eq: auc}
\min_{\x \in \bbR^{n_\x}}\max_{y \in \bbR}f(\x, y) := \frac{1}{N} \sum_{i=1}^{N} f_i(\x, y) + \frac{\lambda}{2} \|\x\|^2 - p(1-p)y^2,
\end{align} 
where $\x = [\w; u; v]^\top \in \bbR^{n_\x}, y \in \bbR$, $\lambda$ is the regularization parameter and $p$ denotes the proportion of positive instances in the dataset. The objective function $f_i(\x, y)$ for each sample is defined as
\begin{align}
\label{eq: auc2}
f_i(\x,y) = &(1-p)\big((\w^\top \a_j-u)^2 - 2(1+y)\w^\top \a_j\big)\I_{b_j=1} \notag\\
&+p\big((\w^\top \a_j-v)^2 + 2(1+y)\w^\top \a_j\big)\I_{b_j=-1}, 
\end{align}
where $\a_i \in \bbR^{n_\x-2}$ are features and $b_i \in \{+1, -1\}$ is the label. Several numerical algorithms have been proposed to solve strongly-convex-strongly-concave minimax optimization problems. In Section~\ref{subsec: AUC/FO}, we introduce first-order algorithms, and in Section~\ref{subsec: AUC/SO}, we discuss a second-order framework for addressing such problems. In this paper, we only consider a linear classifier. However, this approach can be extensively applied to nonlinear models to achieve a better performance such as nonlinear models~\citep{jin2017zeroing, wei2024collaborative, liufu2024neural}  by replacing $\w$ with $h(\w; \a_j)$.

\subsection{First-Order Algorithms}
\label{subsec: AUC/FO}

\subsubsection{Alternating Gradient Descent Ascent}
\label{subsubsec: AUC/FO/GDA}
A commonly used method for solving minimax problems is \textit{Gradient Descent Ascent (GDA)}, which is a generalization of gradient descent from convex optimization~\citep{lin2020gradient, liufu2022modified, huang2023adagda}. In GDA, the parameters are updated either simultaneously or alternately, giving rise to two variations known as \textit{Simultaneous Gradient Descent Ascent (Sim-GDA)} and \textit{Alternating Gradient Descent Ascent (Alt-GDA)}~\citep{zhang2022near}. In Sim-GDA, the parameters are updated simultaneously as follows:
\begin{align*}
\x_{t+1} &= \x_t - \eta \nabla_\x f(\x_t, \y_t), \\
\y_{t+1} &= \y_t + \eta \nabla_\y f(\x_t, \y_t),
\end{align*}
where $\eta$ is the step size. On the other hand, Alt-GDA updates the parameters sequentially, which can lead to more stable convergence, especially in cases where Sim-GDA might oscillate or diverge. The update rules for Alt-GDA are as follows:
\begin{align*}
\x_{t+1} &= \x_t - \eta \nabla_\x f(\x_t, \y_t), \\
\y_{t+1} &= \y_t + \eta \nabla_\y f(\x_{t+1}, \y_t).
\end{align*}
For strongly convex-strongly concave functions, given the precision $\epsilon$, Alt-GDA achieves a convergence rate of $O(\kappa \log(1/\epsilon))$, which is faster than the typical $O(\kappa^2 \log(1/\epsilon))$ rate of Sim-GDA~\citep{zhang2022near}. This implies that Alt-GDA is less sensitive to the condition number $\kappa$ compared to Sim-GDA, which might fail to converge even for general smooth functions.

\subsubsection{ExtraGradient (EG)}
\label{subsubsec: AUC/FO/EG}
The ExtraGradient (EG) method enhances the standard GDA approach by introducing a two-step update mechanism~\citep{mokhtari2020unified}. The main idea of the EG method is to use the gradient at the current point to find an intermediate mid-point:
\begin{align*}
\x_{k+1/2} &= \x_k - \eta \nabla_\x f(\x_k, \y_k), \\
\y_{k+1/2} &= \y_k + \eta \nabla_\y f(\x_k, \y_k).
\end{align*}
Subsequently, the gradients at this mid-point are used to compute the next iteration:
\begin{align*}
\x_{k+1} &= \x_k - \eta \nabla_\x f(\x_{k+1/2}, \y_{k+1/2}), \\
\y_{k+1} &= \y_k + \eta \nabla_\y f(\x_{k+1/2}, \y_{k+1/2}).
\end{align*}
For strongly convex-strongly concave (SCSC) functions, the EG method achieves a convergence rate of $O(\kappa \log(1/\epsilon))$, which is near-optimal for SCSC minimax problems~\citep{mokhtari2020unified}.

\subsection{Second-Order Algorithms}
\label{subsec: AUC/SO}
Second order algorithms are well-studied in convex optimization due to their lower iteration complexity and faster convergence speed. Traditionally, the Newton's method is represented as
\begin{align*}
\begin{bmatrix}
\x_{k+1} \\
\y_{k+1}
\end{bmatrix} = 
\begin{bmatrix}
\x_k \\
\y_k
\end{bmatrix} - 
\underbrace{\left(\begin{bmatrix}
\nabla^2_{\x\x}f(\x_k, \y_k) & \nabla^2_{\x\y}f(\x_k, \y_k) \\
\nabla^2_{\y\x}f(\x_k, \y_k) & \nabla^2_{\y\y}f(\x_k, \y_k)
\end{bmatrix}\right)^{-1}}_{\text{denoted as } \hat{\H}_k^{-1}}
\underbrace{\begin{bmatrix}
\nabla_\x f(\x_k, \y_k) \\
\nabla_\y f(\x_k, \y_k)
\end{bmatrix}}_{\text{denoted as } \g_k},
\end{align*}
which achieves quadratic local convergence with a computational complexity of $\calO(d^3)$ due to the computation of the inverse Hessian matrix, where $d$ is the dimension of the Hessian matrix. Quasi-Newton methods alleviate this burden by approximating the inverse of the Hessian matrix iterative instead of direct computation~\citep{nocedal2006quasi, hennig2013quasi}. However, quasi-Newton methods generally require the Hessian to be positive definite, which is impractical for saddle point problems. A novel reformulation proposed by \cite{liu2022quasi} modifies Newton's method as follows:
\begin{align}
\label{eq:quasi1}
\begin{bmatrix}
\x_{k+1} \\
\y_{k+1}
\end{bmatrix} = 
\begin{bmatrix}
\x_{k} \\
\y_{k}
\end{bmatrix} - \underbrace{\left(\hat{\H}_k^2\right)^{-1}}_{\text{denoted as } \H_k^{-1}} \hat{\H}_k \g_k =
\begin{bmatrix}
\x_{k} \\
\y_{k}
\end{bmatrix} - \H_k^{-1} \hat{\H}_k \g_k,
\end{align}
where $\H_k = \hat{\H}_k^2$ is constructed to be positive definite, which can be approximated by another positive definite matrix $\Q_k$ such that $\H_k \preceq \Q_k$ and hence
\begin{align}\label{eq:quasi2}
\begin{bmatrix}
\x_{k+1} \\
\y_{k+1}
\end{bmatrix} &\approx \begin{bmatrix}
\x_{k} \\
\y_{k}
\end{bmatrix} - \Q_k^{-1} \hat{\H}_k \g_k.
\end{align}
Among quasi-Newton methods, the Broyden family update is the most widely adopted, representing a convex combination of the DFP and SR1 updates as denoted:
\begin{align*}
\text{Broyd}_\tau(\Q, \H, \u) = \tau \text{DFP}(\Q, \H, \u) + (1 - \tau) \text{SR1}(\Q, \H, \u),
\end{align*}
where the parameter $\tau$, which ranges from 0 to 1, balances the contributions of the Symmetric Rank-1 (SR1) and Davidon-Fletcher-Powell (DFP) methods. The SR1 method, updating $\Q$ via a rank-one modification, brings $\Q$ closer in alignment with $\H$ as follows:
\begin{align*}
\text{SR1}(\Q, \H, \u) = \Q - \frac{(\Q - \H)\u\u^\top(\Q-\H)}{\langle (\Q-\H)\u, \u \rangle},
\end{align*}
where $\langle \cdot \rangle$ denotes the inner product of two vectors. The DFP method integrates both current and historical curvature information to achieve a comprehensive approximation:
\begin{align*}
\text{DFP}(\Q, \H, \u) = \Q - &\frac{\H\u\u^\top \Q + \Q\u\u^\top \H}{\langle \H\u, \u\rangle} + \left(1 + \frac{\langle \Q\u, \u\rangle}{\langle \H\u, \u\rangle}\right)\frac{\H\u\u^\top\H}{\langle \H\u, \u \rangle}.
\end{align*}
Specifically, setting $\tau = \frac{\u^\top \H\u}{\u^\top \Q\u}$ leads to the Broyden-Fletcher-Goldfarb-Shanno (BFGS) update:
\begin{align*}
\text{BFGS}(\Q, \H, \u) = \Q - \frac{\Q\u\u^\top \Q}{\langle \Q\u, \u\rangle} + \frac{\H\u\u^\top \H}{\langle \H\u, \u\rangle}.
\end{align*}
Recent advancements include the MGSR1 method proposed by ~\cite{xiao2024multiple}, which enhances performance by integrating greedy selection methods ~\citep{rodomanov2021greedy}. Unlike traditional approaches that randomly select the update vector $\u$ from a standard normal distribution $\calN(0, 1)$ as in \cite{liu2022quasi}, MGSR1 employs a greedy algorithm to select $\u$ by maximizing the ratio:
\begin{align*}
\u_\H(\Q) = \argmax_{\u \in \{\e_1, \ldots, \e_n\}} \frac{\u^\top \Q\u}{\u^\top \H\u},
\end{align*}
The selected $\u$ is then used in the SR1 update, improving the precision of the Hessian's inverse approximation. Additionally, MGSR1 performs multiple greedy selections and updates per iteration, resulting in a more accurate Hessian approximation.

\section{Numerical Experiments}
\label{sec: Experiment}
The numerical experiment was conducted using four incremental feature sets, as detailed in Table~\ref{tab:feature-sets}. The training and test datasets were split following an 80-20 ratio, with standardized preprocessing. To evaluate the model's performance on imbalanced data, we employed several metrics, including accuracy, precision, recall, F1-score and AUC, which are defined as follows:
\begin{gather*}
\text{Accuracy} = \frac{\text{TP} + \text{TN}}{\text{TP} + \text{TN} + \text{FP} + \text{FN}}, \\
\text{Precision} = \frac{\text{TP}}{\text{TP} + \text{FP}}, \\
\text{Recall} = \frac{\text{TP}}{\text{TP} + \text{FN}}, \\
\text{F1-Score} = 2 \times \frac{\text{Precision} \times \text{Recall}}{\text{Precision} + \text{Recall}}.
\end{gather*}
In medical and neuroscience applications, recall is crucial for ensuring that all relevant events, such as seizures or neural abnormalities, are detected, minimizing the risk of missing critical conditions. The F1-score balances recall with precision, making it essential for reliable diagnostics, especially in scenarios with imbalanced data, where both detection accuracy and completeness are vital. The experiments were conducted on Ubuntu 22.04 system, with Python 3.11 and an Nvidia RTX 4080 GPU. For the logistic regression (Section~\ref{subsubsec: Experiment/Baseline/LR}) and linear SVM model (Section~\ref{subsubsec: Experiment/Baseline/SVM}), we tuned the trade-off parameter `C' to achieve best performance. For the numerical algorithms, the termination criterion is the gradient norm threshold such that $\|[\nabla_\x f(\x, y); \nabla_y f(\x, y)]\| \leq 10^{-3}$, or reach the maximum 50,000 iterations.

\begin{table}[htbp]
\centering
\resizebox{0.85\textwidth}{!}{
\begin{tabular}{ccc}
\hline
\addlinespace[2pt]
\textbf{Set} & \textbf{Features Included} & \textbf{Number of Features} \\
\hline
\addlinespace[2pt]
Set 1 & PSD + DE & 112 \\
\hline 
\addlinespace[2pt]
Set 2 & Set 1 + Segment Range + Segment Moments + Segment Diff & 1,008 \\
\hline
\addlinespace[2pt]
Set 3 & Set 2 + Channel Range + Channel Moments & 1,232 \\
\hline
\addlinespace[2pt]
Set 4 & Set 3 + PLV + Corr & 2,792 \\
\hline
\end{tabular}
}
\caption{Feature sets for numerical experiments. Definitions can be found in Table~\ref{tab: feature engineering}.}
\label{tab:feature-sets}
\end{table}

\subsection{Linear Baselines}
\label{subsec: Experiment/Baseline}

\subsubsection{Logistic Regression}
\label{subsubsec: Experiment/Baseline/LR}
Consider a dataset $\{\x_i, y_i\}_{i=1}^{N}$, where $\x_i \in \mathbb{R}^{p}$ are the input predictors and $y_i \in \{1, -1\}$ are the corresponding class labels. Logistic regression is a widely used linear model for binary classification. The log-odds, or logit function, is defined as
\begin{align*}
\log \left(\frac{P(y=1|\mathbf{x}_i)}{1-P(y=1|\mathbf{x}_i)}\right) = \boldsymbol{\beta}^\top \mathbf{x}_i,
\end{align*}
where $P(y=1|\mathbf{x}_i)$ represents the probability that the label $y=1$ given the predictors $\mathbf{x}_i = [1, x_{i1}, x_{i2}, \dots, x_{ip}]^\top \in \mathbb{R}^{p+1}$. Here, $\boldsymbol{\beta} = [\beta_0, \beta_1, \dots, \beta_p]^\top \in \mathbb{R}^{p+1}$ is the vector of coefficients. The logistic function, which maps the linear combination $\boldsymbol{\beta}^\top \mathbf{x}_i$ to a probability value, is given by
\begin{align*}
P(y=1|\mathbf{x}_i) = \frac{1}{1 + \exp(-\boldsymbol{\beta}^\top \mathbf{x}_i)}.
\end{align*}
The model predicts $y = 1$ if $P(y=1 | \mathbf{x}_i) > 0.5$, and $y = -1$ otherwise. This function ensures that the predicted probability is always between 0 and 1, providing a measure of the likelihood of the classification.

\subsubsection{Support Vector Machine}
\label{subsubsec: Experiment/Baseline/SVM}
Given a dataset $\{\mathbf{x}_i, y_i\}_{i=1}^{N}$, where $\mathbf{x}_i \in \mathbb{R}^p$ are the input predictors and $y_i \in \{-1, 1\}$ are the corresponding class labels, the SVM seeks to construct a hyperplane that almost separates the classes by using a soft margin. The optimal hyperplane is determined by solving the following optimization problem:
\begin{align*}
\underset{\boldsymbol{\beta}, \boldsymbol{\epsilon}, M}{\text{maximize}} \qquad & M \\
\text{s.t.} \qquad & \sum_{j=1}^p \beta_j = 1, \\
& y_i \cdot \boldsymbol{\beta}^\top \mathbf{x}_i \geq M(1 - \gamma_i), \quad \forall i = 1, \dots, N \\
& \gamma_i \geq 0, \quad \forall i = 1, \dots, N \quad \text{and} \quad \sum_{i=1}^{N} \gamma_i \leq C,
\end{align*}
where $M$ represents the margin of the hyperplane, $\gamma_i$ are the slack variables that allow for misclassification, $C$ is the regularization parameter controlling the trade-off between maximizing the margin and minimizing the classification error, and $\boldsymbol{\beta} \in \mathbb{R}^{p}$ is the vector of coefficients defining the hyperplane. This optimization problem can be reformulated as a constrained convex optimization problem and solved using methods such as Sequential Minimal Optimization (SMO) or Interior-Point methods. SVM can also be extended to nonlinear classifiers by incorporating kernel methods, which implicitly map the input features into a higher-dimensional space where a linear separation is feasible.

\subsection{Results \& Analysis}
\label{subsec: Experiment/Results}
\begin{figure}[htbp]
\begin{tabular}{ccc}
\includegraphics[scale=0.3]{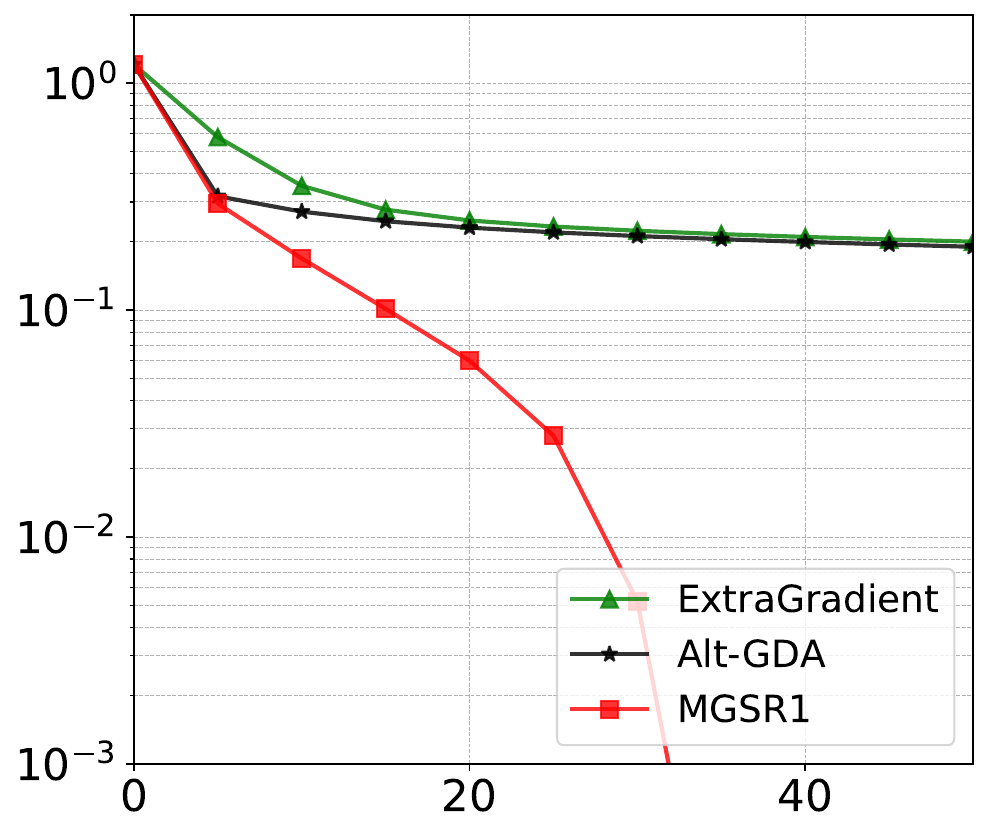} &
\includegraphics[scale=0.3]{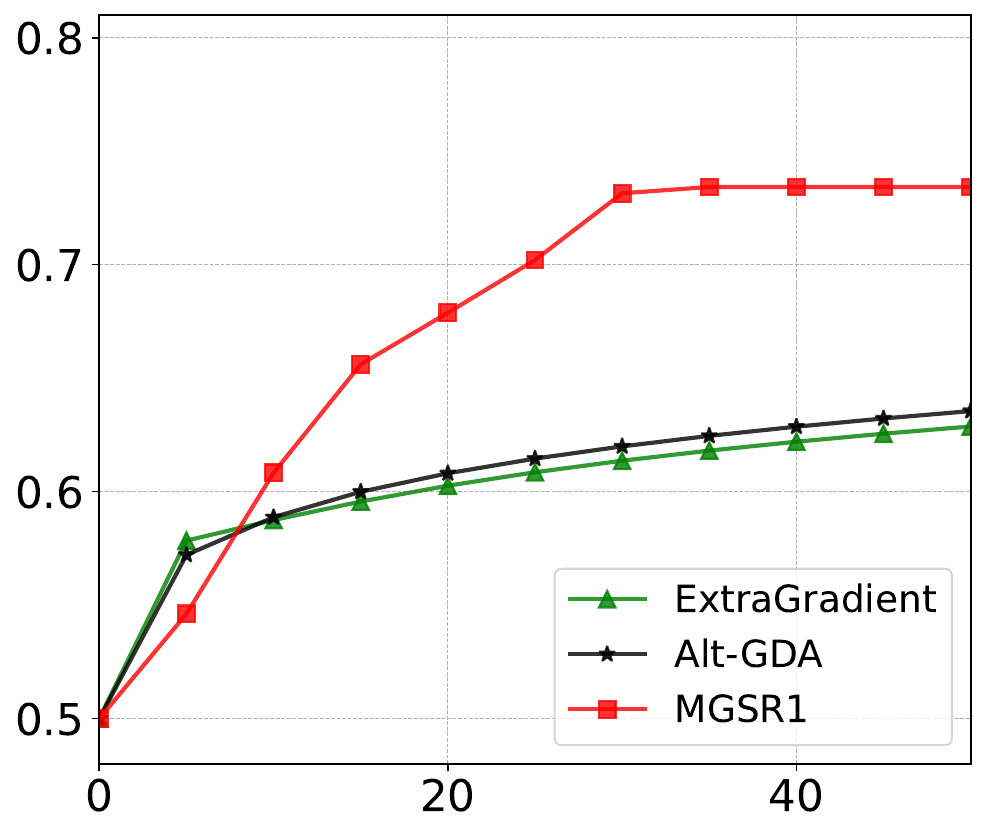} &
\includegraphics[scale=0.3]{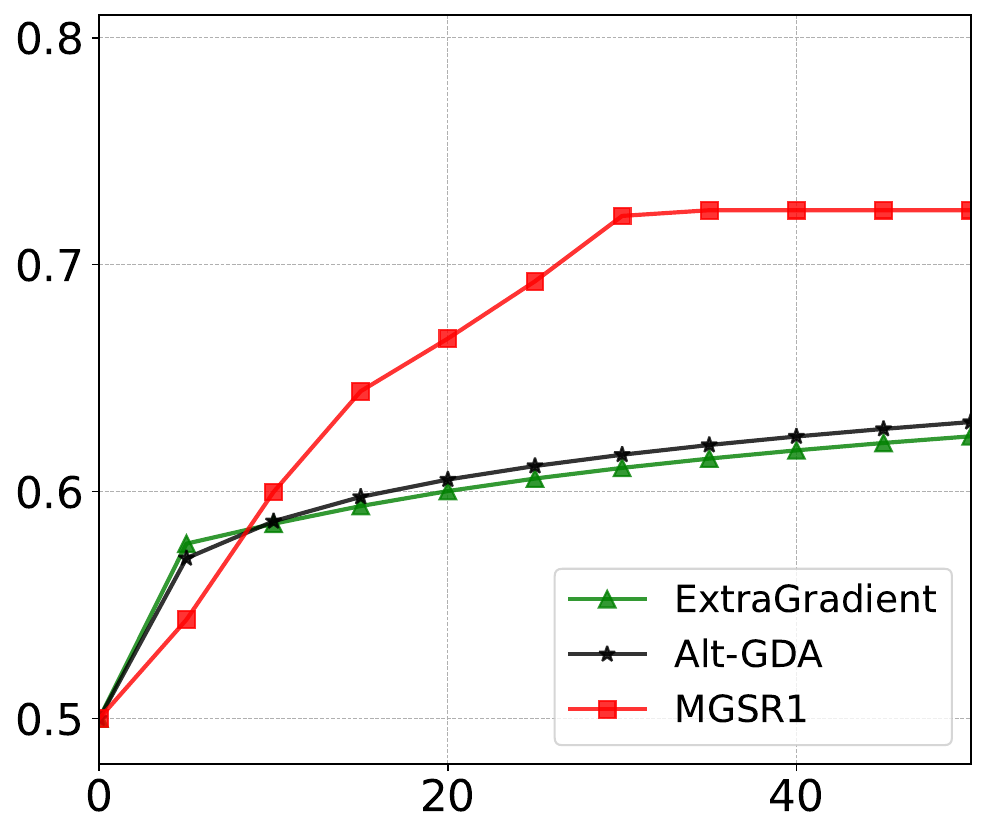} \\
(a) \small{\#Iteration} & (b) \small{\#Iteration} & (c) \small{\#Iteration}
\end{tabular}
\caption{A comparison of numerical optimization algorithms (EG, Alt-GDA, MGSR1) on feature set 3. The y-axis in Fig. (a) represents the gradient norm $\|\nabla f\|_2$, measuring convergence. Figs. (b) and (c) display the Train / Test AUC versus number of iterations.}
\label{fig: converge}
\end{figure}

Figure~\ref{fig: converge} illustrates the training procedures of three numerical algorithms (ExtraGradient, Alt-GDA, and MGSR1) on feature set 3. From an iteration complexity perspective, the second-order algorithm MGSR1 outperforms the first-order algorithms, achieving faster reduction in gradient norm and a quicker increase in AUC score. However, with feature set 4, the second-order algorithm becomes nearly impractical to train due to the computational burden of the large Hessian matrix. Consequently, the numerical optimization results presented in Table~\ref{tab: model result} are obtained using first-order algorithms.

\begin{figure}[htbp]
\begin{tabular}{ccc}
\includegraphics[scale=0.47]{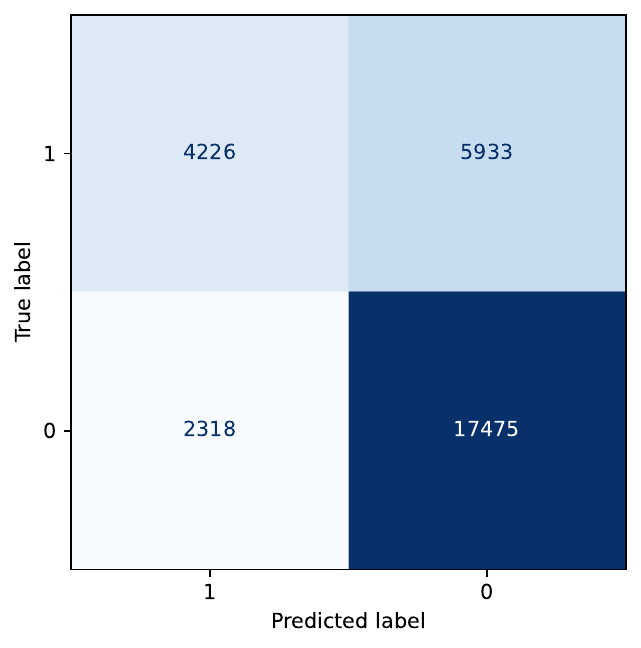} &
\includegraphics[scale=0.47]{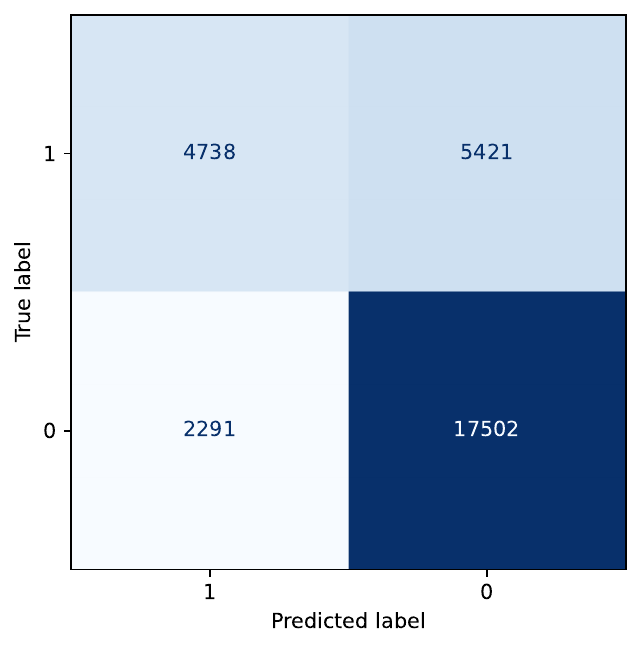} &
\includegraphics[scale=0.47]{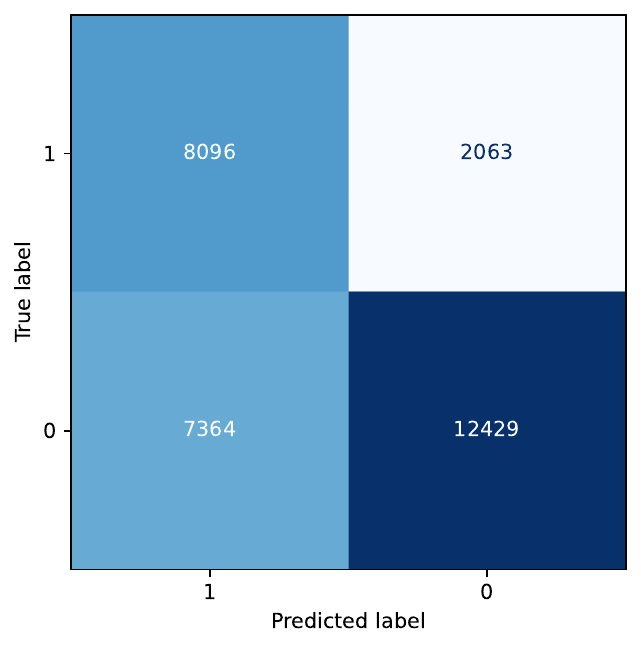} \\[-0.1cm]
~~~(a) \small{Logistic Regression} & ~~~~(b) \small{Linear SVM} & ~~~(c) \small{Numerical}
\end{tabular}
\caption{Confusion matrices for Logistic Regression, Linear SVM, and Numerical Optimization.}
\label{fig: confusion}
\end{figure}

\begin{table}[htbp]
\centering
\captionsetup{font=small, labelfont=bf, labelsep=period}
\caption{Performance comparison across models on train/test sets, with Recall, F1-score, and AUC as key metrics in handling the imbalanced dataset.}
\begin{tabularx}{\textwidth}{@{}ccCCC@{}}
\toprule
 & & \textbf{Logistic Regression} & \textbf{Linear SVM} & \textbf{Numerical} \\
\cmidrule(lr){3-5}
 & & Train / Test & Train / Test & Train / Test \\
\midrule
\multirow{4}{*}{\textbf{Accuracy(\%)}} & Set 1 & 66.74 / 66.63 &  \textbf{67.44} / \textbf{67.25} & 59.08 / 58.81 \\
 & Set 2 & 67.27 / 67.34 &  \textbf{68.05} /  \textbf{67.74} & 60.10 / 59.35 \\
 & Set 3 & 70.88 / 70.34 &  \textbf{72.81} /  \textbf{72.24} & 65.41 / 64.93 \\
 & Set 4 & 74.05 / 72.45 &  \textbf{75.78} /  \textbf{74.25} & 69.94 / 68.53 \\
\addlinespace
\multirow{4}{*}{\textbf{Precision(\%)}} & Set 1 & 51.33 / 53.19 & \textbf{55.45} / \textbf{56.47} & 42.56 / 42.64 \\
 & Set 2 & 56.10 / 54.30 & \textbf{59.41} / \textbf{55.63} & 43.82 / 42.74 \\
 & Set 3 & 63.00 / 61.19 & \textbf{66.18} / \textbf{64.79} & 48.96 / 48.33 \\
 & Set 4 & 67.14 / 64.57 & \textbf{69.56} / \textbf{67.41} & 53.26 / 52.37 \\
\addlinespace
\multirow{4}{*}{\textbf{Recall(\%)}} & Set 1 & 10.58 / 10.59 & 13.41 / 13.48 & \textbf{63.91} / \textbf{63.37} \\
 & Set 2 & 11.63 / 11.27 & 15.34 / 14.59 & \textbf{66.60} / \textbf{63.37} \\
 & Set 3 & 31.99 / 30.62 & 38.71 / 37.03 & \textbf{73.35} / \textbf{71.99} \\
 & Set 4 & 43.73 / 41.60 & 48.92 / 46.64 & \textbf{81.87} / \textbf{79.69} \\
\addlinespace
\multirow{4}{*}{\textbf{F1 score}} & Set 1 & 0.175 / 0.177 & 0.216 / 0.218 & \textbf{0.511} / \textbf{0.510} \\
 & Set 2 & 0.193 / 0.187 & 0.244 / 0.231 & \textbf{0.529} / \textbf{0.517} \\
 & Set 3 & 0.424 / 0.408 & 0.488 / 0.471 & \textbf{0.587} / \textbf{0.578} \\
 & Set 4 & 0.529 / 0.506 & 0.574 / 0.551 &  \textbf{0.645} / \textbf{0.632}\\
\addlinespace
\multirow{4}{*}{\textbf{AUC}} & Set 1 & 0.632 / 0.632 & 0.642 / 0.644 & \textbf{0.645} / \textbf{0.644} \\
 & Set 2 & 0.652 / 0.639 & 0.661 / 0.647 & \textbf{0.664} / \textbf{0.651} \\
 & Set 3 & 0.719 / 0.709 & 0.743 / 0.732 & \textbf{0.744} / \textbf{0.734} \\
 & Set 4 & 0.782 / 0.763 & 0.807 / 0.787 & \textbf{0.812} / \textbf{0.791} \\
\bottomrule
\end{tabularx}
\label{tab: model result}
\end{table}

Figure~\ref{fig: confusion} presents the confusion matrices from three methods trained on feature set 4 (the ultimate feature set), while Table~\ref{tab: model result} provides various performance metrics for these models across four different feature sets, including both training and test sets. The `Liking' label in the DEAP dataset is notably imbalanced, as highlighted in Figure~\ref{fig: labels}. Due to this imbalance, metrics like accuracy and precision tend to be high across all models, but these metrics do not adequately reflect the models' effectiveness in detecting the minority class. The logistic regression and linear SVM models achieved relatively high accuracy and precision, primarily because these metrics are heavily influenced by the overwhelming presence of the majority class (label = 0). However, these traditional metrics fail to capture the models' ability to correctly identify the minority class, leading to a misleading interpretation of model performance.

In contrast, more informative metrics in this imbalanced scenario are recall, F1-score, and AUC. As shown in Table~\ref{tab: model result}, our proposed numerical method, which directly optimizes for AUC, consistently outperformed logistic regression and SVM in these areas. For instance, with feature set 4, the numerical method achieved a recall of 81.87\% on the training set and 79.69\% on the test set, compared to significantly lower recall scores for logistic regression and SVM. The F1-score and AUC metrics similarly indicate superior performance by the numerical method, with the highest AUC of 0.812 on the training set and 0.791 on the test set.

These results demonstrate that the numerical method is particularly effective at identifying the minority class, which is crucial in fields such as neuroscience, and medical diagnostics. In these domains, the ability to accurately detect rare but significant events or conditions (represented by the minority class) can lead to more reliable and robust predictive models, enhancing the overall utility of the system.

\section{Acknowledgment}

Some sentences in Section 4.1 of this manuscript were polished using ChatGPT to improve clarity and readability. The authors take full responsibility for the content of this manuscript, including the refined sections.

\section{Conclusion}
\label{sec: Conclusion}
Imbalanced datasets are a prevalent issue in fields such as cognitive science, neuroscience, and medical diagnostics, where accurately detecting minority classes is critical for reliable model performance. We adopted the DEAP dataset, focusing on the `Liking' label, which we found to be imbalanced—a challenge that has received limited attention in prior research, which predominantly emphasized accuracy metrics. To address this imbalance, we have introduced numerical optimization methods aimed at maximizing the AUC, specifically targeting the improvement of minority class detection. Our approach was tested against traditional models including logistic regression and SVM, showing a substantial enhancement in performance. Specifically, our method achieved an increase in recall from 41.6\% to 79.7\% and improved the F1-score from 0.506 to 0.632, compared to the baseline models. These results underscore the effectiveness of our approach in handling imbalanced data, making it a valuable contribution to the development of robust predictive models in critical areas such as medical diagnostics and neuroscience.

\bibliographystyle{unsrt}
\bibliography{reference.bib}

\end{document}